\documentclass[11pt,a4paper]{article}
\usepackage[hyperref]{acl2018}
\usepackage{times}
\usepackage{latexsym}
\usepackage{amsmath}
\usepackage{amssymb}
\usepackage{extarrows}
\usepackage{multirow}
\usepackage{booktabs}
\usepackage{hyperref}
\usepackage{graphicx}
\makeatletter
\def\UrlAlphabet{%
      \do\a\do\b\do\c\do\d\do\e\do\f\do\g\do\h\do\i\do\j%
      \do\k\do\l\do\m\do\n\do\o\do\p\do\q\do\r\do\s\do\t%
      \do\u\do\v\do\w\do\x\do\y\do\z\do\A\do\B\do\C\do\D%
      \do\E\do\F\do\G\do\H\do\I\do\J\do\K\do\L\do\M\do\N%
      \do\O\do\P\do\Q\do\R\do\S\do\T\do\U\do\V\do\W\do\X%
      \do\Y\do\Z}
\def\UrlDigits{\do\1\do\2\do\3\do\4\do\5\do\6\do\7\do\8\do\9\do\0}
\g@addto@macro{\UrlBreaks}{\UrlOrds}
\g@addto@macro{\UrlBreaks}{\UrlAlphabet}
\g@addto@macro{\UrlBreaks}{\UrlDigits}
\makeatother

\aclfinalcopy 

\newcommand{\tc}[1]{\multicolumn{1}{c}{#1}}

\title{Improving Knowledge Graph Embedding Using Simple Constraints}

\author{Boyang Ding$^{1,2}$, Quan Wang$^{1,2,3}$\thanks{\hspace{0.15cm}Corresponding author: Quan Wang.} , Bin Wang$^{1,2}$, Li Guo$^{1,2}$ \\
  $^{1}$Institute of Information Engineering, Chinese Academy of Sciences\\
  $^{2}$School of Cyber Security, University of Chinese Academy of Sciences\\
  $^{3}$State Key Laboratory of Information Security, Chinese Academy of Sciences\\
  {\tt \{dingboyang,wangquan,wangbin,guoli\}@iie.ac.cn}}

\date{}

\begin{document}
\maketitle
\begin{abstract}
  Embedding knowledge graphs (KGs) into continuous vector spaces is a focus of current research. Early works performed this task via simple models developed over KG triples. Recent attempts focused on either designing more complicated triple scoring models, or incorporating extra information beyond triples. This paper, by contrast, investigates the potential of using very simple constraints to improve KG embedding. We examine {\it non-negativity constraints} on entity representations and {\it approximate entailment constraints} on relation representations. The former help to learn compact and interpretable representations for entities. The latter further encode regularities of logical entailment between relations into their distributed representations. These constraints impose prior beliefs upon the structure of the embedding space, without negative impacts on efficiency or scalability. Evaluation on WordNet, Freebase, and DBpedia shows that our approach is simple yet surprisingly effective, significantly and consistently outperforming competitive baselines. The constraints imposed indeed improve model interpretability, leading to a substantially increased structuring of the embedding space. Code and data are available at \url{https://github.com/iieir-km/ComplEx-NNE_AER}.
\end{abstract}

\section{Introduction}
The past decade has witnessed great achievements in building web-scale knowledge graphs (KGs), {\it e.g.}, Freebase~\cite{bollacker2008:FreeBase}, DBpedia \cite{lehmann2014:DBpedia}, and Google's Knowledge Vault~\cite{dong2014:KnowledgeVault}. A typical KG is a multi-relational graph composed of entities as nodes and relations as different types of edges, where each edge is represented as a triple of the form ({\it head entity}, {\it relation}, {\it tail entity}). Such KGs contain rich structured knowledge, and have proven useful for many NLP tasks~\cite{wasserman2015:WSD,hoffmann2011:IE-Freebase,yang2017:KBLSTM}.

Recently, the concept of {\it knowledge graph embedding} has been presented and quickly become a hot research topic. The key idea there is to embed components of a KG ({\it i.e.}, entities and relations) into a continuous vector space, so as to simplify manipulation while preserving the inherent structure of the KG. Early works on this topic learned such vectorial representations ({\it i.e.}, embeddings) via just simple models developed over KG triples \cite{bordes2011:SE,bordes2013:TransE,jenatton2012:LFM,nickel2011:RESCAL}. Recent attempts focused on either designing more complicated triple scoring models
\cite{socher2013:NTN,bordes2014:SME,wang2014:TransH,lin2015:TransR,xiao2016:TransG,nickel2016:Hole,trouillon2016:ComplEx,liu2017:ANALOGY}, or incorporating extra information beyond KG triples \cite{chang2014:TRESCAL,zhong2015:text,lin2015:PTransE,neelakantan2015:CompositionalVSM,guo2015:SSE,luo2015:path,xie2016:DKRL,xie2016:TKRL,xiao2017:SSP}. See \cite{wang2017:review} for a thorough review.

This paper, by contrast, investigates the potential of using very simple constraints to improve the KG embedding task. Specifically, we examine two types of constraints: (i) {\it non-negativity constraints} on entity representations and (ii) {\it approximate entailment constraints} over relation representations. By using the former, we learn compact representations for entities, which would naturally induce sparsity and interpretability~\cite{murphy2012:NNSE}. By using the latter, we further encode regularities of logical entailment between relations into their distributed representations, which might be advantageous to downstream tasks like link prediction and relation extraction \cite{rocktaschel2015:EmbedLogic,guo2016:KALE}. These constraints impose prior beliefs upon the structure of the embedding space, and will help us to learn more predictive embeddings, without significantly increasing the space or time complexity.

Our work has some similarities to those which integrate logical background knowledge into KG embedding~\cite{rocktaschel2015:EmbedLogic,wang2015:ERInfer,guo2016:KALE,guo2018:RUGE}. Most of such works, however, need grounding of first-order logic rules. The grounding process could be time and space inefficient especially for complicated rules. To avoid grounding, \citet{demeester2016:LiftedRule} tried to model rules using only relation representations. But their work creates vector representations for entity pairs rather than individual entities, and hence fails to handle unpaired entities. Moreover, it can only incorporate strict, hard rules which usually require extensive manual effort to create. \citet{minervini2017:ASR} proposed adversarial training which can integrate first-order logic rules without grounding. But their work, again, focuses on strict, hard rules. \citet{minervini2017:EquivalenceInversion} tried to handle uncertainty of rules. But their work assigns to different rules a same confidence level, and considers only equivalence and inversion of relations, which might not always be available in a given KG.

Our approach differs from the aforementioned works in that: (i) it imposes constraints directly on entity and relation representations without grounding, and can easily scale up to large KGs; (ii) the constraints, {\it i.e.}, non-negativity and approximate entailment derived automatically from statistical properties, are quite universal, requiring no manual effort and applicable to almost all KGs; (iii) it learns an individual representation for each entity, and can successfully make predictions between unpaired entities.

We evaluate our approach on publicly available KGs of WordNet, Freebase, and DBpedia as well. Experimental results indicate that our approach is simple yet surprisingly effective, achieving significant and consistent improvements over competitive baselines, but without negative impacts on efficiency or scalability. The non-negativity and approximate entailment constraints indeed improve model interpretability, resulting in a substantially increased structuring of the embedding space.

The remainder of this paper is organized as follows. We first review related work in Section~\ref{sec:RelatedWork}, and then detail our approach in Section~\ref{sec:Approach}. Experiments and results are reported in Section~\ref{sec:Experiments}, followed by concluding remarks in Section~\ref{sec:Conclusion}.

\section{Related Work}\label{sec:RelatedWork}
Recent years have seen growing interest in learning distributed representations for entities and relations in KGs, a.k.a. KG embedding. Early works on this topic devised very simple models to learn such distributed representations, solely on the basis of triples observed in a given KG, {\it e.g.}, TransE which takes relations as translating operations between head and tail entities \cite{bordes2013:TransE}, and RESCAL which models triples through bilinear operations over entity and relation representations \cite{nickel2011:RESCAL}. Later attempts roughly fell into two groups: (i) those which tried to design more complicated triple scoring models, {\it e.g.}, the TransE extensions~\cite{wang2014:TransH,lin2015:TransR,ji2015:TransD}, the RESCAL extensions \cite{yang2015:Bilinear,nickel2016:Hole,trouillon2016:ComplEx,liu2017:ANALOGY}, and the (deep) neural network models \cite{socher2013:NTN,bordes2014:SME,shi2017:ProjE,schlichtkrull2017:R-GCN,dettmers2017:ConvE}; (ii) those which tried to integrate extra information beyond triples, {\it e.g.}, entity types~\cite{guo2015:SSE,xie2016:TKRL}, relation paths \cite{neelakantan2015:CompositionalVSM,lin2015:PTransE}, and textual descriptions \cite{xie2016:DKRL,xiao2017:SSP}. Please refer to \cite{nickel2016:review,wang2017:review} for a thorough review of these techniques. In this paper, we show the potential of using very simple constraints ({\it i.e.}, non-negativity constraints and approximate entailment constraints) to improve KG embedding, without significantly increasing the model complexity.

A line of research related to ours is KG embedding with logical background knowledge incorporated \cite{rocktaschel2015:EmbedLogic,wang2015:ERInfer,guo2016:KALE,guo2018:RUGE}. But most of such works require grounding of first-order logic rules, which is time and space inefficient especially for complicated rules. To avoid grounding, \citet{demeester2016:LiftedRule} proposed lifted rule injection, and \citet{minervini2017:ASR} investigated adversarial training. Both works, however, can only handle strict, hard rules which usually require extensive effort to create. \citet{minervini2017:EquivalenceInversion} tried to handle uncertainty of background knowledge. But their work considers only equivalence and inversion between relations, which might not always be available in a given KG. Our approach, in contrast, imposes constraints directly on entity and relation representations without grounding. And the constraints used are quite universal, requiring no manual effort and applicable to almost all KGs.

Non-negativity has long been a subject studied in various research fields. Previous studies reveal that non-negativity could naturally induce sparsity and, in most cases, better interpretability \cite{lee1999:NMF}. In many NLP-related tasks, non-negativity constraints are introduced to learn more interpretable word representations, which capture the notion of semantic composition \cite{murphy2012:NNSE,luo2015:NN,fyshe2015:NN}. In this paper, we investigate the ability of non-negativity constraints to learn more accurate KG embeddings with good interpretability.

\section{Our Approach}\label{sec:Approach}
This section presents our approach. We first introduce a basic embedding technique to model triples in a given KG (\S~\ref{sebsec:ComplEx}). Then we discuss the non-negativity constraints over entity representations (\S~\ref{sebsec:Non-negativity}) and the approximate entailment constraints over relation representations (\S~\ref{subsec:Entailment}). And finally we present the overall model (\S~\ref{subsec:OverallModel}).

\subsection{A Basic Embedding Model}\label{sebsec:ComplEx}
We choose ComplEx~\cite{trouillon2016:ComplEx} as our basic embedding model, since it is simple and efficient, achieving state-of-the-art predictive performance. Specifically, suppose we are given a KG containing a set of triples $\mathcal{O}=\{(e_i,r_k,e_j)\}$, with each triple composed of two entities $e_i, e_j \in \mathcal{E}$ and their relation $r_k \in \mathcal{R}$. Here $\mathcal{E}$ is the set of entities and $\mathcal{R}$ the set of relations. ComplEx then represents each entity $e \in \mathcal{E}$ as a complex-valued vector $\mathbf{e}$ $\in \mathbb{C}^d$, and each relation $r \in \mathcal{R}$ a complex-valued vector $\mathbf{r} \in \mathbb{C}^d$, where $d$ is the dimensionality of the embedding space. Each $\mathbf{x} \in \mathbb{C}^d$ consists of a real vector component $\textrm{Re}(\mathbf{x})$ and an imaginary vector component $\textrm{Im}(\mathbf{x})$, {\it i.e.}, $\mathbf{x}=\textrm{Re}(\mathbf{x}) + i\textrm{Im}(\mathbf{x})$. For any given triple $(e_i,r_k,e_j) \in \mathcal{E} \times \mathcal{R} \times \mathcal{E}$, a multi-linear dot product is used to score that triple, {\it i.e.},
\begin{align}\label{eq:ComplEx}
\phi(e_i,r_k,e_j) &\triangleq \textrm{Re}(\langle \mathbf{e}_i, \mathbf{r}_k, \bar{\mathbf{e}}_j\rangle) \notag \\
                  &\triangleq \textrm{Re}(\sum\nolimits_\ell [\mathbf{e}_i]_\ell [\mathbf{r}_k]_\ell [\bar{\mathbf{e}}_j]_\ell),
\end{align}
where $\mathbf{e}_i, \mathbf{r}_k, \mathbf{e}_j \in \mathbb{C}^d$ are the vectorial representations associated with $e_i, r_k, e_j$, respectively; $\bar{\mathbf{e}}_j$ is the conjugate of $\mathbf{e}_j$; $[\cdot]_\ell$ is the $\ell$-th entry of a vector; and $\textrm{Re}(\cdot)$ means taking the real part of a complex value. Triples with higher $\phi(\cdot,\cdot,\cdot)$ scores are more likely to be true. Owing to the asymmetry of this scoring function, {\it i.e.}, $\phi(e_i,r_k,e_j) \neq \phi(e_j,r_k,e_i)$, ComplEx can effectively handle asymmetric relations~\cite{trouillon2016:ComplEx}.

\subsection{Non-negativity of Entity Representations}\label{sebsec:Non-negativity}
On top of the basic ComplEx model, we further require entities to have non-negative (and bounded) vectorial representations. In fact, these distributed representations can be taken as feature vectors for entities, with latent semantics encoded in different dimensions. In ComplEx, as well as most (if not all) previous approaches, there is no limitation on the range of such feature values, which means that both positive and negative properties of an entity can be encoded in its representation. However, as pointed out by \citet{murphy2012:NNSE}, it would be uneconomical to store all negative properties of an entity or a concept. For instance, to describe cats (a concept), people usually use positive properties such as cats are mammals, cats eat fishes, and cats have four legs, but hardly ever negative properties like cats are not vehicles, cats do not have wheels, or cats are not used for communication.

Based on such intuition, this paper proposes to impose non-negativity constraints on entity representations, by using which only positive properties will be stored in these representations. To better compare different entities on the same scale, we further require entity representations to stay within the hypercube of $[0,1]^d$, as approximately Boolean embeddings~\cite{kruszewski2015:BooleanEmbedding}, {\it i.e.},
\begin{equation}\label{eq:Non-negativity}
\mathbf{0} \leq \textrm{Re}(\mathbf{e}), \textrm{Im}(\mathbf{e}) \leq \mathbf{1}, \quad \forall e \in \mathcal{E},
\end{equation}
where $\mathbf{e}\in\mathbb{C}^d$ is the representation for entity $e\in\mathcal{E}$, with its real and imaginary components denoted by $\textrm{Re}(\mathbf{e}), \textrm{Im}(\mathbf{e}) \in \mathbb{R}^d$; $\mathbf{0}$ and $\mathbf{1}$ are $d$-dimensional vectors with all their entries being $0$ or $1$; and $\geq, \leq, =$ denote the entry-wise comparisons throughout the paper whenever applicable. As shown by~\citet{lee1999:NMF}, non-negativity, in most cases, will further induce sparsity and interpretability.

\subsection{Approximate Entailment for Relations}\label{subsec:Entailment}
Besides the non-negativity constraints over entity representations, we also study approximate entailment constraints over relation representations. By approximate entailment, we mean an ordered pair of relations that the former approximately entails the latter, {\it e.g.}, \texttt{\small BornInCountry} and \texttt{\small Nationality}, stating that a person born in a country is very likely, but not necessarily, to have a nationality of that country. Each such relation pair is associated with a weight to indicate the confidence level of entailment. A larger weight stands for a higher level of confidence. We denote by $r_p \xrightarrow{\lambda} r_q$ the approximate entailment between relations $r_p$ and $r_q$, with confidence level $\lambda$. This kind of entailment can be derived automatically from a KG by modern rule mining systems \cite{galarraga2015:AMIE+}. Let $\mathcal{T}$ denote the set of all such approximate entailments derived beforehand.

Before diving into approximate entailment, we first explore the modeling of strict entailment, {\it i.e.}, entailment with infinite confidence level $\lambda=+\infty$. The strict entailment $r_p \rightarrow r_q$ states that if relation $r_p$ holds then relation $r_q$ must also hold. This entailment can be roughly modelled by requiring
\begin{equation}\label{eq:Implication}
\phi(e_i,r_p,e_j) \leq \phi(e_i,r_q,e_j), \quad \forall e_i,e_j\in\mathcal{E},
\end{equation}
where $\phi(\cdot,\cdot,\cdot)$ is the score for a triple predicted by the embedding model, defined by Eq.~(\ref{eq:ComplEx}). Eq.~(\ref{eq:Implication}) can be interpreted as follows: for any two entities $e_i$ and $e_j$, if $(e_i,r_p,e_j)$ is a true fact with a high score $\phi(e_i,r_p,e_j)$, then the triple $(e_i,r_q,e_j)$ with an even higher score should also be predicted as a true fact by the embedding model. Note that given the non-negativity constraints defined by Eq.~(\ref{eq:Non-negativity}), a sufficient condition for Eq.~(\ref{eq:Implication}) to hold, is to further impose
\begin{equation}\label{eq:StrictEntailment}
\textrm{Re}(\mathbf{r}_p) \leq \textrm{Re}(\mathbf{r}_q), \;\; \textrm{Im}(\mathbf{r}_p) = \textrm{Im}(\mathbf{r}_q),
\end{equation}
where $\mathbf{r}_p$ and $\mathbf{r}_q$ are the complex-valued representations for $r_p$ and $r_q$ respectively, with the real and imaginary components denoted by $\textrm{Re}(\cdot),\textrm{Im}(\cdot) \in \mathbb{R}^d$. That means, when the constraints of Eq.~(\ref{eq:StrictEntailment}) (along with those of Eq.~(\ref{eq:Non-negativity})) are satisfied, the requirement of Eq.~(\ref{eq:Implication}) (or in other words $r_p \rightarrow r_q$) will always hold. We provide a proof of sufficiency as Appendix~\ref{subsec:Sufficiency}.

Next we examine the modeling of approximate entailment. To this end, we further introduce the confidence level $\lambda$ and allow slackness in Eq.~(\ref{eq:StrictEntailment}), which yields
\begin{align}
\lambda \big(\textrm{Re}(\mathbf{r}_p) - \textrm{Re}(\mathbf{r}_q)\big) \leq \boldsymbol{\alpha}, \label{eq:ApproximateEntailment-1} \\
\lambda \big(\textrm{Im}(\mathbf{r}_p) - \textrm{Im}(\mathbf{r}_q)\big)^2 \leq \boldsymbol{\beta}. \label{eq:ApproximateEntailment-2}
\end{align}
Here $\boldsymbol{\alpha}, \boldsymbol{\beta} \geq \mathbf{0}$ are slack variables, and $(\cdot)^2$ means an entry-wise operation. Entailments with higher confidence levels show less tolerance for violating the constraints. When $\lambda=+\infty$, Eqs.~(\ref{eq:ApproximateEntailment-1})~--~(\ref{eq:ApproximateEntailment-2}) degenerate to Eq.~(\ref{eq:StrictEntailment}). The above analysis indicates that our approach can model entailment simply by imposing constraints over relation representations, without traversing all possible $(e_i, e_j)$ entity pairs ({\it i.e.}, grounding). In addition, different confidence levels are encoded in the constraints, making our approach moderately tolerant of uncertainty.

\subsection{The Overall Model}\label{subsec:OverallModel}
Finally, we combine together the basic embedding model of ComplEx, the non-negativity constraints on entity representations, and the approximate entailment constraints over relation representations. The overall model is presented as follows:
\begin{align}\label{eq:ConstrainedKGE}
\min_{\Theta, \{\boldsymbol{\alpha}, \boldsymbol{\beta}\}} \;\; & \sum_{\mathcal{D}^+\cup\mathcal{D}^-} \!\! \log \big( 1 + \exp (\!-y_{ijk} \phi(e_i,r_k,e_j)) \big) \notag \\
& + \mu \sum\nolimits_{\mathcal{T}} \boldsymbol{1}^\top (\boldsymbol{\alpha} + \boldsymbol{\beta}) + \eta \|\Theta\|_2^2, \notag \\
\textrm{s.t.} \;\; & \lambda \big(\textrm{Re}(\mathbf{r}_p) - \textrm{Re}(\mathbf{r}_q)\big) \leq \boldsymbol{\alpha}, \notag \\
              \;\; & \lambda \big(\textrm{Im}(\mathbf{r}_p) - \textrm{Im}(\mathbf{r}_q)\big)^2 \leq \boldsymbol{\beta}, \notag \\
              \;\; & \boldsymbol{\alpha}, \boldsymbol{\beta} \geq \mathbf{0}, \quad \forall r_p \xrightarrow{\lambda} r_q \in \mathcal{T}, \notag \\
              \;\; & \mathbf{0} \leq \textrm{Re}(\mathbf{e}), \textrm{Im}(\mathbf{e}) \leq \mathbf{1}, \quad \forall e \in \mathcal{E}.
\end{align}
Here, $\Theta \triangleq \{\mathbf{e}: e\in\mathcal{E}\}\cup\{\mathbf{r}: r\in\mathcal{R}\}$ is the set of all entity and relation representations; $\mathcal{D}^+$ and $\mathcal{D}^-$ are the sets of positive and negative training triples respectively; a positive triple is directly observed in the KG, {\it i.e.}, $(e_i,r_k,e_j)\in\mathcal{O}$; a negative triple can be generated by randomly corrupting the head or the tail entity of a positive triple, {\it i.e.}, $(e_i',r_k,e_j)$ or $(e_i,r_k,e_j')$; $y_{ijk}=\pm1$ is the label (positive or negative) of triple $(e_i,r_k,e_j)$. In this optimization, the first term of the objective function is a typical logistic loss, which enforces triples to have scores close to their labels. The second term is the sum of slack variables in the approximate entailment constraints, with a penalty coefficient $\mu \geq 0$. The motivation is, although we allow slackness in those constraints we hope the total slackness to be small, so that the constraints can be better satisfied. The last term is $L_2$ regularization to avoid over-fitting, and $\eta \geq 0$ is the regularization coefficient.

To solve this optimization problem, the approximate entailment constraints (as well as the corresponding slack variables) are converted into penalty terms and added to the objective function, while the non-negativity constraints remain as they are. As such, the optimization problem of Eq.~(\ref{eq:ConstrainedKGE}) can be rewritten as:
\begin{align}\label{eq:RegularizedKGE}
\min_\Theta \;\; & \sum_{\mathcal{D}^+\cup\mathcal{D}^-} \!\! \log \big( 1 + \exp (\!-y_{ijk} \phi(e_i,r_k,e_j)) \big) \notag \\
& + \mu \!\sum\nolimits_{\mathcal{T}} \lambda \boldsymbol{1}^\top \!\big[\textrm{Re}(\mathbf{r}_p) \!-\! \textrm{Re}(\mathbf{r}_q)\big]_+ \notag \\
& + \mu \!\sum\nolimits_{\mathcal{T}} \lambda \boldsymbol{1}^\top \!\big(\textrm{Im}(\mathbf{r}_p) \!-\! \textrm{Im}(\mathbf{r}_q)\big)^2 \notag \!+ \eta \|\Theta\|_2^2, \notag \\
\textrm{s.t.} \;\; & \mathbf{0} \leq \textrm{Re}(\mathbf{e}), \textrm{Im}(\mathbf{e}) \leq \mathbf{1}, \quad \forall e \in \mathcal{E},
\end{align}
where $[\mathbf{x}]_+ = \max(\mathbf{0}, \mathbf{x})$ with $\max(\cdot,\cdot)$ being an entry-wise operation. The equivalence between Eq.~(\ref{eq:ConstrainedKGE}) and Eq.~(\ref{eq:RegularizedKGE}) is shown in the Appendix~\ref{subsec:Equivalence}. We use SGD in mini-batch mode as our optimizer, with AdaGrad~\cite{duchi2011:AdaGrad} to tune the learning rate. After each gradient descent step, we project (by truncation) real and imaginary components of entity representations into the hypercube of $[0,1]^d$, to satisfy the non-negativity constraints.

While favouring a better structuring of the embedding space, imposing the additional constraints will not substantially increase model complexity. Our approach has a space complexity of $O(nd+md)$, which is the same as that of ComplEx. Here, $n$ is the number of entities, $m$ the number of relations, and $O(nd+md)$ to store a $d$-dimensional complex-valued vector for each entity and each relation. The time complexity (per iteration) of our approach is $O(sd+td+\bar{n}d)$, where $s$ is the average number of triples in a mini-batch, $\bar{n}$ the average number of entities in a mini-batch, and $t$ the total number of approximate entailments in $\mathcal{T}$. $O(sd)$ is to handle triples in a mini-batch, $O(td)$ penalty terms introduced by the approximate entailments, and $O(\bar{n}d)$ further the non-negativity constraints on entity representations. Usually there are much fewer entailments than triples, {\it i.e.}, $t \ll s$, and also $\bar{n} \leq 2s$.\footnote{There will be at most $2s$ entities contained in $s$ triples.} So the time complexity of our approach is on a par with $O(sd)$, {\it i.e.}, the time complexity of ComplEx.

\section{Experiments and Results}\label{sec:Experiments}
This section presents our experiments and results. We first introduce the datasets used in our experiments (\S~\ref{subsec:Datasets}). Then we empirically evaluate our approach in the link prediction task (\S~\ref{subsec:LinkPrediction}). After that, we conduct extensive analysis on both entity representations (\S~\ref{subsec:EntityRepresentations}) and relation representations (\S~\ref{subsec:RelationRepresentations}) to show the interpretability of our model. Code and data used in the experiments are available at \url{https://github.com/iieir-km/ComplEx-NNE_AER}.

\subsection{Datasets}\label{subsec:Datasets}
The first two datasets we used are WN18 and FB15K, released by \citet{bordes2013:TransE}.\footnote{\url{https://everest.hds.utc.fr/doku.php?id=en:smemlj12}} WN18 is a subset of WordNet containing 18 relations and 40,943 entities, and FB15K a subset of Freebase containing 1,345 relations and 14,951 entities. We create our third dataset from the mapping-based objects of core DBpedia.\footnote{\url{http://downloads.dbpedia.org/2016-10/core/}} We eliminate relations not included within the DBpedia ontology such as \texttt{\small HomePage} and \texttt{\small Logo}, and discard entities appearing less than 20 times. The final dataset, referred to as DB100K, is composed of 470 relations and 99,604 entities. Triples on each datasets are further divided into training, validation, and test sets, used for model training, hyperparameter tuning, and evaluation respectively. We follow the original split for WN18 and FB15K, and draw a split of 597,572/ 50,000/50,000 triples for DB100K.

We further use AMIE+ \cite{galarraga2015:AMIE+}\footnote{\url{https://www.mpi-inf.mpg.de/departments/databases-and-information-systems/research/yago-naga/amie/}} to extract approximate entailments automatically from the \textit{training} set of each dataset. As suggested by \citet{guo2018:RUGE}, we consider entailments with PCA confidence higher than 0.8.\footnote{PCA confidence is the confidence under the partial completeness assumption. See \cite{galarraga2015:AMIE+} for details.} As such, we extract 17 approximate entailments from WN18, 535 from FB15K, and 56 from DB100K. Table~\ref{tab:Rules} gives some examples of these approximate entailments, along with their confidence levels. Table~\ref{tab:Dataset} further summarizes the statistics of the datasets.

\begin{table}[!t]
  \begin{center}\footnotesize\setlength{\tabcolsep}{3pt}
  \begin{tabular*}{0.48 \textwidth}{@{\extracolsep{\fill}}@{}l@{}}
        \toprule
        $\textrm{hypernym}^{-1} \xrightarrow{1.00} \textrm{hyponym}$ \\
        $\textrm{synset\_domain\_topic\_of}^{-1} \xrightarrow{0.99} \textrm{member\_of\_domain\_topic}$ \\
        $\textrm{instance\_hypernym}^{-1} \xrightarrow{0.98} \textrm{instance\_hyponym}$ \\
        \midrule
        $\textrm{/people/place\_of\_birth}^{-1} \xrightarrow{1.00} \textrm{/location/people\_born\_here}$ \\
        $\textrm{/film/directed\_by}^{-1} \xrightarrow{0.98} \textrm{/director/film}$ \\
        $\textrm{/country/admin\_divisions} \xrightarrow{0.91} \textrm{/country/1st\_level\_divisions}$ \\
        \midrule
        $\textrm{owner} \xrightarrow{0.95} \textrm{owning\_company}$                                                  \\
        $\textrm{child}^{-1} \xrightarrow{0.92} \textrm{parent}$    \\
        $\textrm{distributing\_company} \xrightarrow{0.92} \textrm{distributing\_label}$                                                  \\
        \bottomrule
  \end{tabular*}
\end{center}
\caption{\label{tab:Rules} Approximate entailments extracted from WN18 (top), FB15K (middle), and DB100K (bottom), where $r^{-1}$ means the inverse of relation $r$. }
\end{table}

\begin{table}[!t]
    \centering\footnotesize\setlength{\tabcolsep}{3pt}
    \begin{tabular*}{0.48 \textwidth}{@{\extracolsep{\fill}}@{}l|rrrrrr@{}}
        \toprule
        Dataset & \#~Ent & \#~Rel & \multicolumn{3}{c}{\#~Train/Valid/Test} & \#~Cons \\
        \midrule
        WN18    & 40,943 & 18     & 141,442    & 5,000        & 5,000       & 17  \\
        FB15K   & 14,951 & 1,345  & 483,142    & 50,000       & 59,071      & 535 \\
        DB100K  & 99,604 & 470    & 597,572    & 50,000       & 50,000      & 56  \\
        \bottomrule
    \end{tabular*}
    \caption{\label{tab:Dataset} Statistics of datasets, where the columns respectively indicate the number of entities, relations, training/validation/test triples, and approximate entailments.}
\end{table}

\subsection{Link Prediction}\label{subsec:LinkPrediction}
We first evaluate our approach in the link prediction task, which aims to predict a triple $(e_i, r_k, e_j)$ with $e_i$ or $e_j$ missing, {\it i.e.}, predict $e_i$ given $(r_k, e_j)$ or predict $e_j$ given $(e_i, r_k)$.

\smallskip
\textbf{Evaluation Protocol:} We follow the protocol introduced by \citet{bordes2013:TransE}. For each test triple $(e_i, r_k, e_j)$, we replace its head entity $e_i$ with every entity $e_i' \in \mathcal{E}$, and calculate a score for the corrupted triple $(e_i', r_k, e_j)$, {\it e.g.}, $\phi(e_i', r_k, e_j)$ defined by Eq.~(\ref{eq:ComplEx}). Then we sort these scores in descending order, and get the rank of the correct entity $e_i$. During ranking, we remove corrupted triples that already exist in either the training, validation, or test set, {\it i.e.}, the {\it filtered} setting as described in \cite{bordes2013:TransE}. This whole procedure is repeated while replacing the tail entity $e_j$. We report on the {\it test} set the mean reciprocal rank (MRR) and the proportion of correct entities ranked in the top $n$ (HITS@N), with $n=1,3,10$.

\smallskip
\textbf{Comparison Settings:} We compare the performance of our approach against a variety of KG embedding models developed in recent years. These models can be categorized into three groups:
\begin{itemize}
  \item Simple embedding models that utilize triples alone without integrating extra information, including TransE \cite{bordes2013:TransE}, DistMult \cite{yang2015:Bilinear}, HolE \cite{nickel2016:Hole}, ComplEx \cite{trouillon2016:ComplEx}, and ANALOGY \cite{liu2017:ANALOGY}. Our approach is developed on the basis of ComplEx.
  \item Other extensions of ComplEx that integrate logical background knowledge in addition to triples, including RUGE \cite{guo2018:RUGE} and ComplEx$^\textrm{R}$ \cite{minervini2017:EquivalenceInversion}. The former requires grounding of first-order logic rules. The latter is restricted to relation equivalence and inversion, and assigns an identical confidence level to all different rules.
  \item Latest developments or implementations that achieve current state-of-the-art performance reported on the benchmarks of WN18 and FB15K, including R-GCN \cite{schlichtkrull2017:R-GCN}, ConvE \cite{dettmers2017:ConvE}, and Single DistMult \cite{kadlec2017:Baselines}.\footnote{We do not consider Ensemble DistMult \cite{dettmers2017:ConvE} which combines several different models together, to facilitate a fair comparison.} The first two are built based on neural network architectures, which are, by nature, more complicated than the simple models. The last one is a re-implementation of DistMult, generating 1000 to 2000 negative training examples per positive one, which leads to better performance but requires significantly longer training time.
\end{itemize}

We further evaluate our approach in two different settings: (i) ComplEx-NNE that imposes only the \underline{N}on-\underline{N}egativity constraints on \underline{E}ntity representations, {\it i.e.}, optimization Eq.~(\ref{eq:RegularizedKGE}) with $\mu=0$; and (ii) ComplEx-NNE+AER that further imposes the \underline{A}pproximate \underline{E}ntailment constraints over \underline{R}elation representations besides those non-negativity ones, {\it i.e.}, optimization Eq.~(\ref{eq:RegularizedKGE}) with $\mu>0$.

\smallskip
\textbf{Implementation Details:} We compare our approach against all the three groups of baselines on the benchmarks of WN18 and FB15K. We directly report their original results on these two datasets to avoid re-implementation bias. On DB100K, the newly created dataset, we take the first two groups of baselines, {\it i.e.}, those simple embedding models and ComplEx extensions with logical background knowledge incorporated. We do not use the third group of baselines due to efficiency and complexity issues. We use the code provided by \citet{trouillon2016:ComplEx}\footnote{\url{https://github.com/ttrouill/complex}} for TransE, DistMult, and ComplEx, and the code released by their authors for ANALOGY\footnote{\url{https://github.com/quark0/ANALOGY}} and RUGE\footnote{\url{https://github.com/iieir-km/RUGE}}. We re-implement HolE and ComplEx$^\textrm{R}$ so that all the baselines (as well as our approach) share the same optimization mode, {\it i.e.}, SGD with AdaGrad and gradient normalization, to facilitate a fair comparison.\footnote{An exception here is that ANALOGY uses asynchronous SGD with AdaGrad~\cite{liu2017:ANALOGY}.} We follow \citet{trouillon2016:ComplEx} to adopt a ranking loss for TransE and a logistic loss for all the other methods.

\begin{table*}[!t]
    \centering\footnotesize\setlength{\tabcolsep}{5pt}
    \begin{tabular*}{1 \textwidth}{@{\extracolsep{\fill}}@{}llllllllll@{}}
    \toprule
    & \multicolumn{4}{c}{WN18} && \multicolumn{4}{c}{FB15K} \\\cmidrule{2-5}\cmidrule{7-10}
    & & \multicolumn{3}{c}{HITS@N} & & & \multicolumn{3}{c}{HITS@N} \\\cmidrule{3-5}\cmidrule{8-10}
                                                &MRR         &\tc{1}      &\tc{3}      &\tc{10}     &&MRR         &\tc{1}      &\tc{3}      &\tc{10}     \\
    \midrule
    TransE~\cite{bordes2013:TransE}             &0.454       &0.089       &0.823       &0.934       &&0.380       &0.231       &0.472       &0.641       \\
    DistMult~\cite{yang2015:Bilinear}           &0.822       &0.728       &0.914       &0.936       &&0.654       &0.546       &0.733       &0.824       \\
    HolE~\cite{nickel2016:Hole}                 &0.938       &0.930       &0.945       &0.949       &&0.524       &0.402       &0.613       &0.739       \\
    ComplEx~\cite{trouillon2016:ComplEx}        &0.941       &0.936       &0.945       &0.947       &&0.692       &0.599       &0.759       &0.840       \\
    ANALOGY~\cite{liu2017:ANALOGY}              &0.942       &0.939       &0.944       &0.947       &&0.725       &0.646       &0.785       &0.854       \\
    \midrule
    RUGE~\cite{guo2018:RUGE}                    &\tc{---}    &\tc{---}    &\tc{---}    &\tc{---}    &&0.768       &0.703       &0.815       &0.865       \\
    ComplEx$^\textrm{R}$~\cite{minervini2017:EquivalenceInversion}
                                                &0.940       &\tc{---}    &0.943       &0.947       &&\tc{---}    &\tc{---}    &\tc{---}    &\tc{---}    \\
    \midrule
    R-GCN~\cite{schlichtkrull2017:R-GCN}        &0.814       &0.686       &0.928       &0.955       &&0.651       &0.541       &0.736       &0.825       \\
    R-GCN+~\cite{schlichtkrull2017:R-GCN}       &0.819       &0.697       &0.929       &{\bf 0.964} &&0.696       &0.601       &0.760       &0.842       \\
    ConvE~\cite{dettmers2017:ConvE}             &0.942       &0.935       &{\bf 0.947} &0.955       &&0.745       &0.670       &0.801       &0.873       \\
    Single DistMult~\cite{kadlec2017:Baselines} &0.797       &\tc{---}    &\tc{---}    &0.946       &&0.798       &\tc{---}    &\tc{---}    &{\bf 0.893} \\
    \midrule
    ComplEx-NNE (this work)                     &0.941       &0.937       &0.944       &0.948       &&0.727$^*$       &0.659$^*$       &0.772$^*$       &0.845$^*$       \\
    ComplEx-NNE+AER (this work)                 &{\bf 0.943} &{\bf 0.940} &0.945       &0.948       &&{\bf 0.803}$^*$ &{\bf 0.761}$^*$ &{\bf 0.831}$^*$ &0.874$^*$       \\
    \bottomrule
    \end{tabular*}
    \caption{\label{tab:LinkPrediction} Link prediction results on the test sets of WN18 and FB15K. Results for TransE and DistMult are taken from \cite{trouillon2016:ComplEx}. Results for the other baselines are taken from the original papers. Missing scores not reported in the literature are indicated by ``---''. Best scores are highlighted in bold, and ``$*$" indicates statistically significant improvements over ComplEx.}
\end{table*}

\begin{table}[!t]
    \centering\footnotesize\setlength{\tabcolsep}{5pt}
    \begin{tabular*}{0.48 \textwidth}{@{\extracolsep{\fill}}@{}lllll@{}}
    \toprule
    & & \multicolumn{3}{c}{HITS@N} \\\cmidrule{3-5}
    & MRR & \tc{1}    & \tc{3}    & \tc{10}  \\
    \midrule
    TransE               & 0.111       & 0.016       & 0.164       & 0.270       \\
    DistMult             & 0.233       & 0.115       & 0.301       & {\bf 0.448} \\
    HolE                 & 0.260       & 0.182       & 0.309       & 0.411       \\
    ComplEx              & 0.242       & 0.126       & 0.312       & 0.440       \\
    ANALOGY              & 0.252       & 0.143       & 0.323       & 0.427       \\
    \midrule
    RUGE                 & 0.246       & 0.129       & 0.325       & 0.433       \\
    ComplEx$^\textrm{R}$ & 0.253       & 0.167       & 0.294       & 0.420       \\
    \midrule
    ComplEx-NNE          & 0.298$^*$       & 0.229$^*$       & 0.330$^*$       & 0.426       \\
    ComplEx-NNE+AER      & {\bf 0.306}$^*$ & {\bf 0.244}$^*$ & {\bf 0.334}$^*$ & 0.418       \\
    \bottomrule
    \end{tabular*}
    \caption{\label{tab:LinkPrediction-DB100K} Link prediction results on the test set of DB100K, with best scores highlighted in bold, statistically significant improvements marked by ``$*$".}
\end{table}

Among those baselines, RUGE and ComplEx$^\textrm{R}$ require additional logical background knowledge. RUGE makes use of soft rules, which are extracted by AMIE+ from the {\it training} sets. As suggested by \citet{guo2018:RUGE}, length-1 and length-2 rules with PCA confidence higher than 0.8 are utilized. Note that our approach also makes use of AMIE+ rules with PCA confidence higher than 0.8. But it only considers entailments between a pair of relations, {\it i.e.}, length-1 rules. ComplEx$^\textrm{R}$ takes into account equivalence and inversion between relations. We derive such axioms directly from our approximate entailments. If $r_p \xrightarrow{\lambda_1} r_q$ and $r_q \xrightarrow{\lambda_2} r_p$ with $\lambda_1,\lambda_2$ $>0.8$, we think relations $r_p$ and $r_q$ are equivalent. And similarly, if $r_p^{-1} \xrightarrow{\lambda_1} r_q$ and $r_q^{-1} \xrightarrow{\lambda_2} r_p$ with $\lambda_1,\lambda_2>0.8$, we consider $r_p$ as an inverse of $r_q$.

For all the methods, we create 100 mini-batches on each dataset, and conduct a grid search to find hyperparameters that maximize MRR on the validation set, with at most 1000 iterations over the training set. Specifically, we tune the embedding size $d \in \{100, 150, 200\}$, the $L_2$ regularization coefficient $\eta \!\in\! \{0.001, 0.003, 0.01, 0.03, 0.1\}$, the ratio of negative over positive training examples $\alpha$ $\in \{ 2, 10\}$, and the initial learning rate $\gamma \in \{0.01,$ $0.05, 0.1, 0.5, 1.0\}$. For TransE, we tune the margin of the ranking loss $\delta \in \{0.1, 0.2, 0.5, 1, 2, 5,$ $10\}$. Other hyperparameters of ANALOGY and RUGE are set or tuned according to the default settings suggested by their authors \cite{liu2017:ANALOGY,guo2018:RUGE}. After getting the best ComplEx model, we tune the relation constraint penalty of our approach ComplEx-NNE+AER ($\mu$ in Eq.~(\ref{eq:RegularizedKGE})) in the range of $\{10^{-5}, 10^{-4}, \cdots, 10^4, 10^5\}$, with all its other hyperparameters fixed to their optimal configurations. We then directly set $\mu=0$ to get the optimal ComplEx-NNE model. The weight of soft constraints in ComplEx$^\textrm{R}$ is tuned in the same range as $\mu$. The optimal configurations for our approach are: $d=200$, $\eta=0.03$, $\alpha=10$, $\gamma=1.0$, $\mu=10$ on WN18; $d=200$, $\eta\!=\!0.01$, $\alpha\!=\!10$, $\gamma=$ $0.5$, $\mu=10^{-3}$ on FB15K; and $d=150$, $\eta=0.03$, $\alpha=10$, $\gamma=0.1$, $\mu=10^{-5}$ on DB100K.

\smallskip
\textbf{Experimental Results:} Table~\ref{tab:LinkPrediction} presents the results on the test sets of WN18 and FB15K, where the results for the baselines are taken directly from previous literature. Table~\ref{tab:LinkPrediction-DB100K} further provides the results on the test set of DB100K, with all the methods tuned and tested in (almost) the same setting. On all the datasets, we test statistical significance of the improvements achieved by ComplEx-NNE/ ComplEx-NNE+AER over ComplEx, by using a paired t-test. The reciprocal rank or HITS@N value with $n=1,3,10$ for each test triple is used as paired data. The symbol ``$*$" indicates a significance level of $p<0.05$.

The results demonstrate that imposing the non-negativity and approximate entailment constraints indeed improves KG embedding. ComplEx-NNE and ComplEx-NNE+AER perform better than (or at least equally well as) ComplEx in almost all the metrics on all the three datasets, and most of the improvements are statistically significant (except those on WN18). More interestingly, just by introducing these simple constraints, ComplEx-NNE+ AER can beat very strong baselines, including the best performing basic models like ANALOGY, those previous extensions of ComplEx like RUGE or ComplEx$^\textrm{R}$, and even the complicated developments or implementations like ConvE or Single DistMult. This demonstrates the superiority of our approach.

\subsection{Analysis on Entity Representations}\label{subsec:EntityRepresentations}
\begin{figure}[!t]
\centering
  \includegraphics[width=0.48\textwidth]{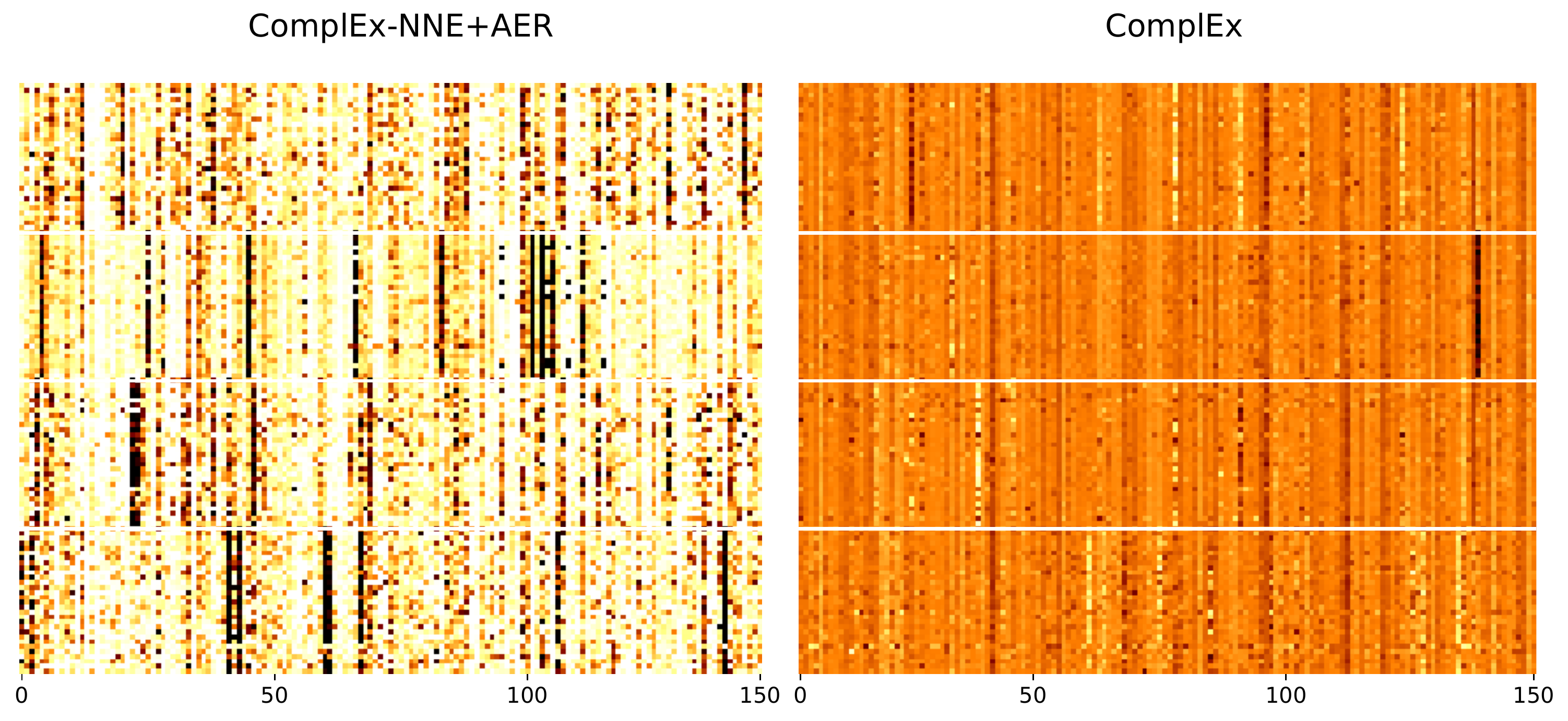}
  \caption{Visualization of real components of entity representations (rows) learned by ComplEx-NNE+AER (left) and ComplEx (right). From top to bottom, entities belong to type \texttt{\small reptile}, \texttt{\small wine\_} \texttt{\small region}, \texttt{\small species}, and \texttt{\small programming\_language} in turn. Values range from 0 (white) via 0.5 (orange) to 1 (black). Best viewed in color.}\label{fig:Entity-Rep}
\end{figure}

This section inspects how the structure of the entity embedding space changes when the constraints are imposed. We first provide the visualization of entity representations on DB100K. On this dataset each entity is associated with a single type label.\footnote{\url{http://downloads.dbpedia.org/2016-10/core-i18n/en/instance_types_wkd_uris_en.ttl.bz2}} We pick 4 types \texttt{\small reptile}, \texttt{\small wine\_region}, \texttt{\small species}, and \texttt{\small programming\_language}, and randomly select 30 entities from each type. Figure~\ref{fig:Entity-Rep} visualizes the representations of these entities learned by ComplEx and ComplEx-NNE+AER (real components only), with the optimal configurations determined by link prediction (see \S~\ref{subsec:LinkPrediction} for details, applicable to all analysis hereafter). During the visualization, we normalize the real component of each entity by $[\tilde{\mathbf{x}}]_\ell \!=\! \frac{[\mathbf{x}]_\ell - \min(\mathbf{x})}{\max(\mathbf{x}) - \min(\mathbf{x})}$, where $\min(\mathbf{x})$ or $\max(\mathbf{x})$ is the minimum or maximum entry of $\mathbf{x}$ respectively. We observe that after imposing the non-negativity constraints, ComplEx-NNE+AER indeed obtains compact and interpretable representations for entities. Each entity is represented by only a relatively small number of ``active'' dimensions. And entities with the same type tend to activate the same set of dimensions, while entities with different types often get clearly different dimensions activated.

\begin{figure}[!t]
\centering
  \includegraphics[width=0.48\textwidth]{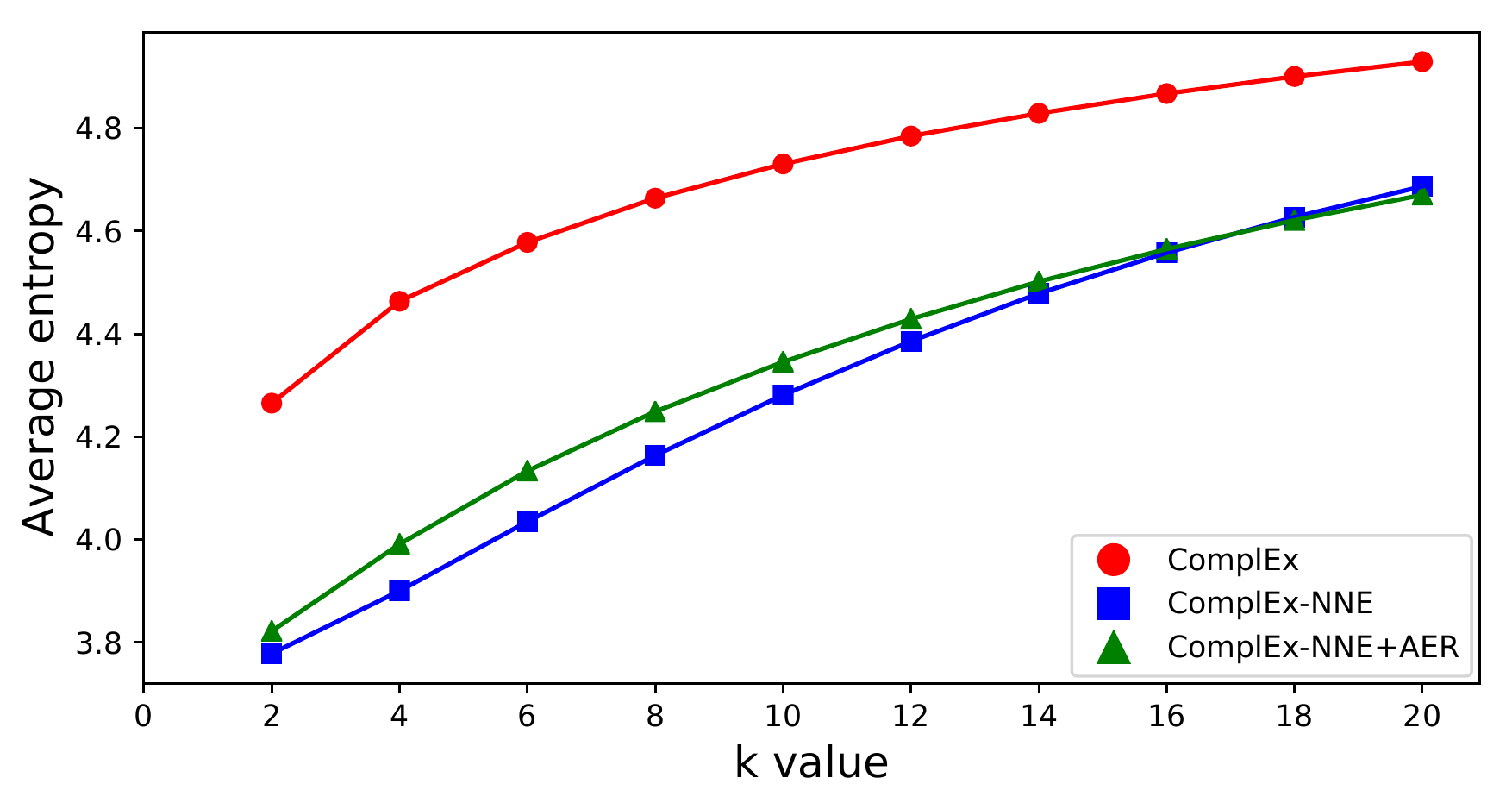}
  \caption{Average entropy over all dimensions of real components of entity representations learned by ComplEx (circles), ComplEx-NNE (squares), and ComplEx-NNE+AER (triangles) as $K$ varies.}\label{fig:Entropy}
\end{figure}

Then we investigate the semantic purity of these dimensions. Specifically, we collect the representations of all the entities on DB100K (real components only). For each dimension of these representations, top $K$ percent of entities with the highest activation values on this dimension are picked. We can calculate the entropy of the type distribution of the entities selected. This entropy reflects diversity of entity types, or in other words, semantic purity. If all the $K$ percent of entities have the same type, we will get the lowest entropy of zero (the highest semantic purity). On the contrary, if each of them has a distinct type, we will get the highest entropy (the lowest semantic purity). Figure~\ref{fig:Entropy} shows the average entropy over all dimensions of entity representations (real components only) learned by ComplEx, ComplEx-NNE, and ComplEx-NNE+ AER, as $K$ varies. We can see that after imposing the non-negativity constraints, ComplEx-NNE and ComplEx-NNE+AER can learn entity representations with latent dimensions of consistently higher semantic purity. We have conducted the same analyses on imaginary components of entity representations, and observed similar phenomena. The results are given as Appendix~\ref{subsec:Imaginary}.

\begin{figure}[!t]
\centering
  \includegraphics[width=0.45\textwidth]{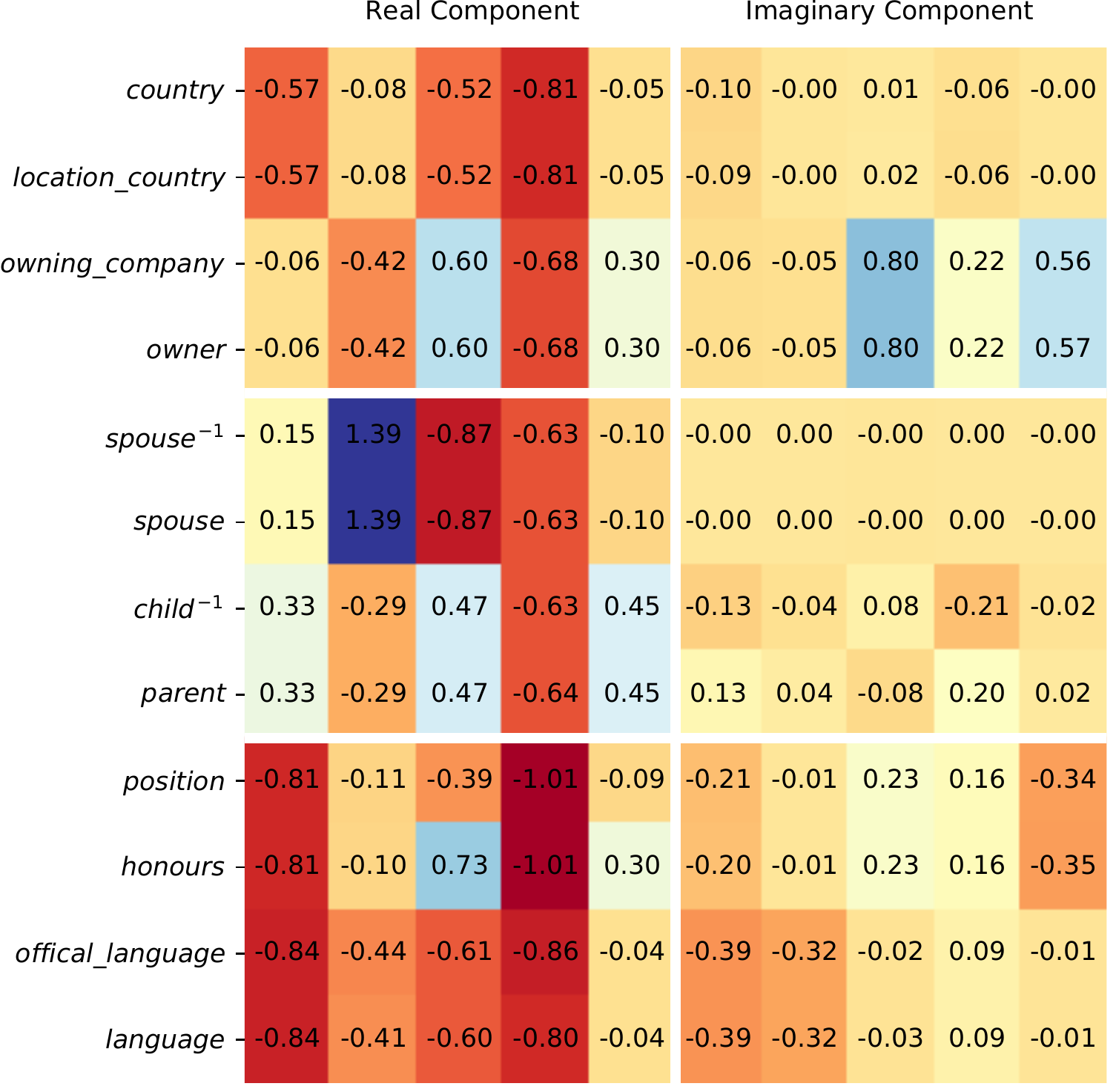}
  \caption{Visualization of relation representations learned by ComplEx-NNE+AER, with the top 4 relations from the equivalence class, the middle 4 the inversion class, and the bottom 4 others.}\label{fig:Relation-Rep}
\end{figure}

\subsection{Analysis on Relation Representations}\label{subsec:RelationRepresentations}
This section further provides a visual inspection of the relation embedding space when the constraints are imposed. To this end, we group relation pairs involved in the DB100K entailment constraints into 3 classes: equivalence, inversion, and others.\footnote{Equivalence and inversion are detected using heuristics introduced in \S~\ref{subsec:LinkPrediction} (implementation details). See the Appendix~\ref{subsec:Properties} for detailed properties of these three classes.} We choose 2 pairs of relations from each class, and visualize these relation representations learned by ComplEx-NNE+AER in Figure~\ref{fig:Relation-Rep}, where for each relation we randomly pick 5 dimensions from both its real and imaginary components. By imposing the approximate entailment constraints, these relation representations can encode logical regularities quite well. Pairs of relations from the first class (equivalence) tend to have identical representations $\mathbf{r}_p \approx \mathbf{r}_q$, those from the second class (inversion) complex conjugate representations $\mathbf{r}_p \approx \bar{\mathbf{r}}_q$; and the others representations that $\textrm{Re}(\mathbf{r}_p) \leq \textrm{Re}(\mathbf{r}_q)$ and $\textrm{Im}(\mathbf{r}_p) \approx \textrm{Im}(\mathbf{r}_q)$.

\section{Conclusion}\label{sec:Conclusion}
This paper investigates the potential of using very simple constraints to improve KG embedding. Two types of constraints have been studied: (i) the non-negativity constraints to learn compact, interpretable entity representations, and (ii) the approximate entailment constraints to further encode logical regularities into relation representations. Such constraints impose prior beliefs upon the structure of the embedding space, and will not significantly increase the space or time complexity. Experimental results on benchmark KGs demonstrate that our method is simple yet surprisingly effective, showing significant and consistent improvements over strong baselines. The constraints indeed improve model interpretability, yielding a substantially increased structuring of the embedding space.

\section*{Acknowledgments}
We would like to thank all the anonymous reviewers for their insightful and valuable suggestions, which help to improve the quality of this paper. This work is supported by the National Key Research and Development Program of China (No. 2016QY03D0503) and the Fundamental Theory and Cutting Edge Technology Research Program of the Institute of Information Engineering, Chinese Academy of Sciences (No. Y7Z0261101).


\appendix

\section{Supplemental Material}\label{sec:Supplemental}

\subsection{Sufficient Condition for Eq.~(\ref{eq:Implication})}\label{subsec:Sufficiency}
Given the non-negativity constraints of Eq.~(\ref{eq:Non-negativity}), a sufficient condition for Eq.~(\ref{eq:Implication}) to hold, is to further impose the strict entailment constraints of Eq.~(\ref{eq:StrictEntailment}). In fact, given the constraints of Eq.~(\ref{eq:Non-negativity}) and Eq.~(\ref{eq:StrictEntailment}), we will always have
\begin{equation*}
\begin{aligned}
\phi(e_i,r_p,e_j) \;=\; & \langle \textrm{Re}(\mathbf{e}_i), \textrm{Re}(\mathbf{r}_p), \textrm{Re}(\mathbf{e}_j) \rangle \\
& + \langle \textrm{Im}(\mathbf{e}_i), \textrm{Re}(\mathbf{r}_p), \textrm{Im}(\mathbf{e}_j) \rangle \\
& + \langle \textrm{Re}(\mathbf{e}_i), \textrm{Im}(\mathbf{r}_p), \textrm{Im}(\mathbf{e}_j) \rangle \\
& - \langle \textrm{Im}(\mathbf{e}_i), \textrm{Im}(\mathbf{r}_p), \textrm{Re}(\mathbf{e}_j) \rangle \\
\;\leq\; & \langle \textrm{Re}(\mathbf{e}_i), \textrm{Re}(\mathbf{r}_q), \textrm{Re}(\mathbf{e}_j) \rangle \\
& + \langle \textrm{Im}(\mathbf{e}_i), \textrm{Re}(\mathbf{r}_q), \textrm{Im}(\mathbf{e}_j) \rangle \\
& + \langle \textrm{Re}(\mathbf{e}_i), \textrm{Im}(\mathbf{r}_q), \textrm{Im}(\mathbf{e}_j) \rangle \\
& - \langle \textrm{Im}(\mathbf{e}_i), \textrm{Im}(\mathbf{r}_q), \textrm{Re}(\mathbf{e}_j) \rangle \\
\;=\; & \phi(e_i,r_q,e_j)
\end{aligned}
\end{equation*}
for any two entities $e_i, e_j \in \mathcal{E}$, {\it i.e.}, Eq.~(\ref{eq:Implication}). Here, the first two terms of the inequality hold because $\textrm{Re}(\mathbf{r}_p) \leq \textrm{Re}(\mathbf{r}_q)$, and the last two terms because $\textrm{Im}(\mathbf{r}_p) = \textrm{Im}(\mathbf{r}_q)$, given the condition that $\textrm{Re}(\mathbf{e}),$ $\textrm{Im}(\mathbf{e}) \geq \mathbf{0}$ for every $e\in\mathcal{E}$.

\subsection{Equivalence between Eq.~(\ref{eq:ConstrainedKGE}) and Eq.~(\ref{eq:RegularizedKGE})}\label{subsec:Equivalence}
We first rewrite the constraints of the optimization Eq.~(\ref{eq:ConstrainedKGE}). Specifically, the two constraints
\begin{equation*}
\boldsymbol{\alpha} \geq \lambda \big(\textrm{Re}(\mathbf{r}_p) - \textrm{Re}(\mathbf{r}_q)\big), \;\; \boldsymbol{\alpha} \geq \boldsymbol{0}
\end{equation*}
can be rewritten as a single one, {\it i.e.},
\begin{equation*}
\boldsymbol{\alpha} \geq \lambda \big[\textrm{Re}(\mathbf{r}_p) - \textrm{Re}(\mathbf{r}_q)\big]_+,
\end{equation*}
where $[\mathbf{x}]_+ = \max(\mathbf{0}, \mathbf{x})$ with $\max(\cdot,\cdot)$ being an entry-wise operator. Similarly, the two constraints
\begin{equation*}
\boldsymbol{\beta} \geq \lambda \big(\textrm{Im}(\mathbf{r}_p) - \textrm{Im}(\mathbf{r}_q)\big)^2, \;\; \boldsymbol{\beta} \geq \boldsymbol{0}
\end{equation*}
degenerate to a single one, {\it i.e.},
\begin{equation*}
\boldsymbol{\beta} \geq \lambda \big(\textrm{Im}(\mathbf{r}_p) - \textrm{Im}(\mathbf{r}_q)\big)^2.
\end{equation*}
As the objective function of Eq.~(\ref{eq:ConstrainedKGE}) has to minimize $\boldsymbol{1}^\top (\boldsymbol{\alpha} + \boldsymbol{\beta})$ over all possible $\boldsymbol{\alpha}, \boldsymbol{\beta}$, an optimal value for this term will be
\begin{equation*}
\lambda \boldsymbol{1}^\top \big[\textrm{Re}(\mathbf{r}_p) - \textrm{Re}(\mathbf{r}_q)\big]_+ + \lambda \boldsymbol{1}^\top \big(\textrm{Im}(\mathbf{r}_p) - \textrm{Im}(\mathbf{r}_q)\big)^2.
\end{equation*}
Plugging this back into the objective function and removing the degenerated constraints, we will obtain the optimization of Eq.~(\ref{eq:RegularizedKGE}).

\subsection{Analyses on Imaginary Components of Entity Representations}\label{subsec:Imaginary}
\begin{figure}[b]
	\centering
	\includegraphics[width=0.48\textwidth]{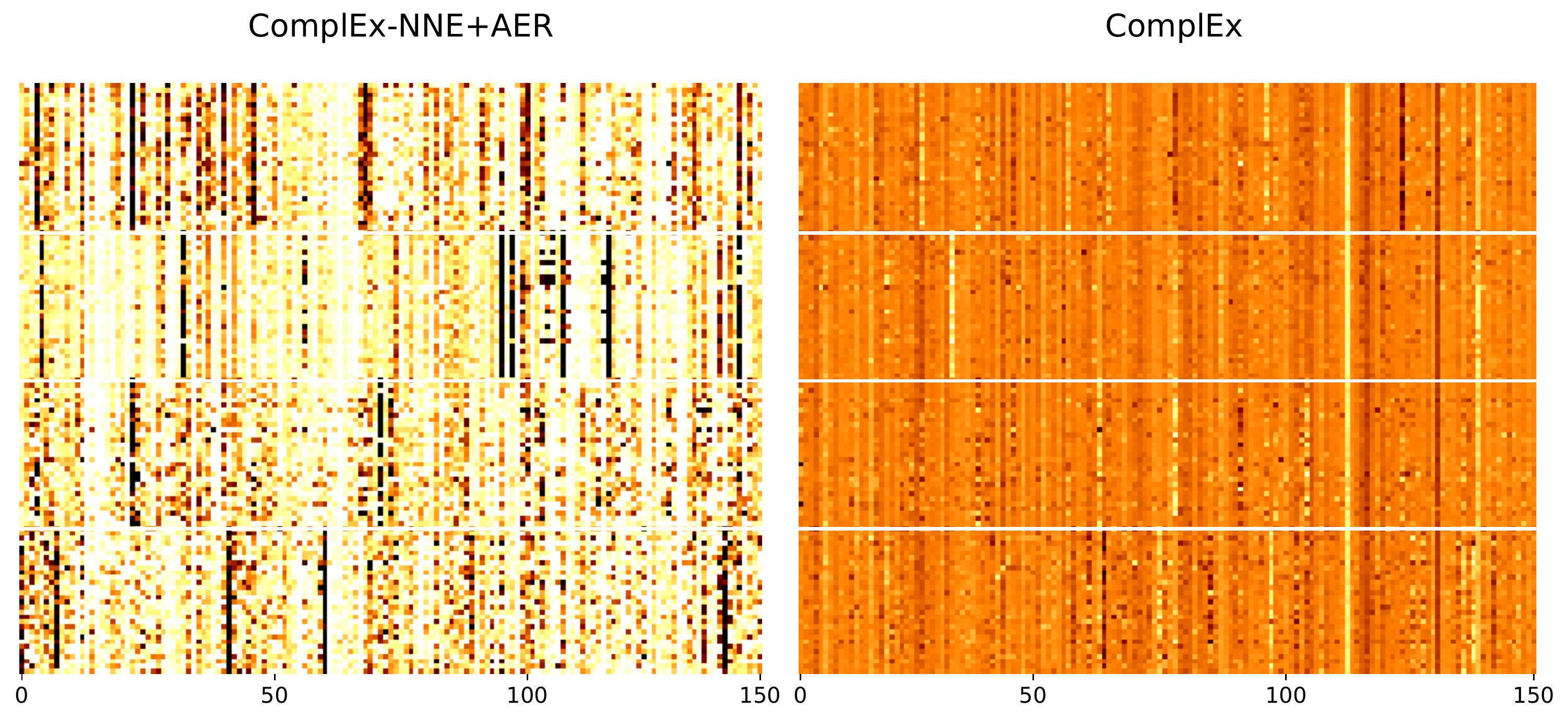}
	\caption{Visualization of imaginary components of entity representations (rows) learned by ComplEx-NNE+AER (left) and ComplEx (right). From top to bottom, entities belong to \texttt{\small reptile}, \texttt{\small wine\_} \texttt{\small region}, \texttt{\small species}, \texttt{\small programming\_language} in turn. Values range from 0 (white) via 0.5 (orange) to 1 (black). Best viewed in color.}\label{fig:Entity-Rep-imag}
\end{figure}

\begin{figure}[t]
	\centering
	\includegraphics[width=0.48\textwidth]{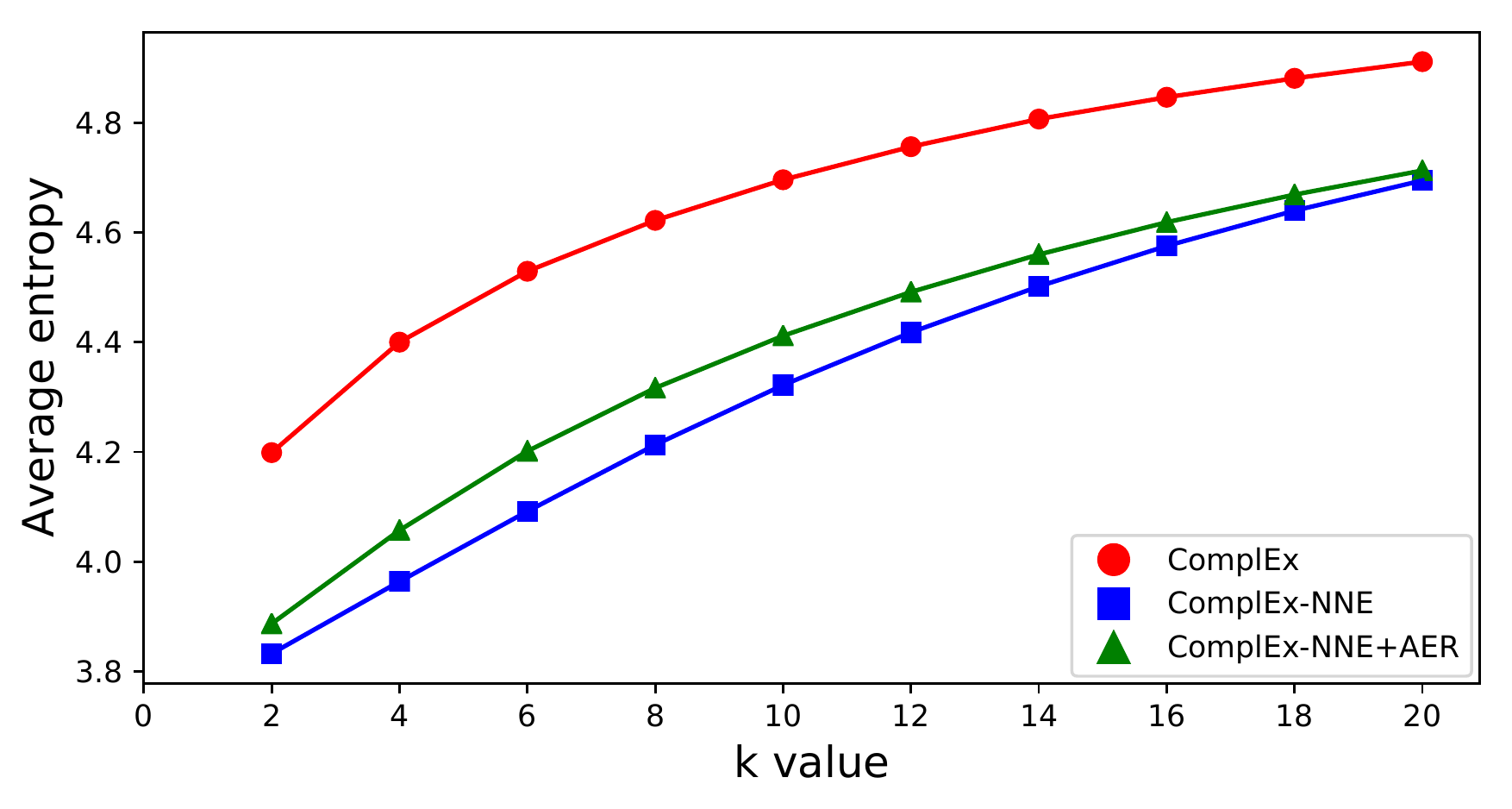}
	\caption{Average entropy over all dimensions of imaginary components of entity representations learned by ComplEx (circles), ComplEx-NNE (squares), and ComplEx-NNE+AER (triangles) as $K$ varies.}\label{fig:Entropy-imag}
\end{figure}

We conduct the same analyses on imaginary components of entity representations as those conducted on real ones (\S~4.3). Figure~\ref{fig:Entity-Rep-imag} visualizes imaginary components of entity representations learned by ComplEx and ComplEx-NNE+AER, with the optimal configurations determined by link prediction. Figure~\ref{fig:Entropy-imag} shows average entropy along imaginary components of entity representations learned by ComplEx, ComplEx-NNE, and ComplEx-NNE +AER.

\subsection{Properties of Equivalence, Inversion, or Ordinary Entailment}\label{subsec:Properties}
For ordinary entailment $r_p \rightarrow r_q$ (neither equivalence nor inversion), the constraints of Eq.~(\ref{eq:StrictEntailment}) directly suggest
\begin{equation*}
\textrm{Re}(\mathbf{r}_p) \leq \textrm{Re}(\mathbf{r}_q), \;\; \textrm{Im}(\mathbf{r}_p) = \textrm{Im}(\mathbf{r}_q).
\end{equation*}
For equivalence $r_p \leftrightarrow r_q$ ($r_p \rightarrow r_q$ and $r_q \rightarrow r_p$), we ought to have
\begin{align*}
\textrm{Re}(\mathbf{r}_p) \leq \textrm{Re}(\mathbf{r}_q), \;\; \textrm{Im}(\mathbf{r}_p) = \textrm{Im}(\mathbf{r}_q), \\
\textrm{Re}(\mathbf{r}_q) \leq \textrm{Re}(\mathbf{r}_p), \;\; \textrm{Im}(\mathbf{r}_q) = \textrm{Im}(\mathbf{r}_p),
\end{align*}
which imply $\mathbf{r}_p \!=\! \mathbf{r}_q$. Since
\begin{align*}
\phi(e_i,r_k,e_j) & = \textrm{Re}(\langle \mathbf{e}_i, \mathbf{r}_k, \bar{\mathbf{e}}_j\rangle) \\
& = \textrm{Re}(\langle \mathbf{e}_j, \bar{\mathbf{r}}_k, \bar{\mathbf{e}}_i\rangle) \\
& \triangleq \phi(e_j,r_k^{-1},e_i)
\end{align*}
for any $e_i,e_j\in\mathcal{E}$ and $r_k\in\mathcal{R}$, we could represent the inverse of relation $r_k$ ({\it i.e.} $r_k^{-1}$) as the conjugate of $\mathbf{r}_k$ ({\it i.e.} $\bar{\mathbf{r}}_k$). Then for inversion $r_p \leftrightarrow r_q^{-1}$, we ought to have $\mathbf{r}_p = \bar{\mathbf{r}}_q$.

\end{document}